\documentclass[12pt,draftcls,onecolumn]{IEEEtran}
                            
\usepackage{makeidx}         
\usepackage{graphicx}        
                                        
\usepackage[pdftex]{color}
\usepackage{multicol}        
\usepackage[bottom]{footmisc} 
\usepackage{subfig}
\usepackage[normalem]{ulem}

\makeindex            
                       
\usepackage{amsfonts,amsmath,amssymb,euscript,mathbbol,dsfont,mathtools,txfonts}

\usepackage{tabularx}  
\usepackage{ragged2e}  
\newcolumntype{Y}{>{\RaggedRight\arraybackslash}X} 
\usepackage{booktabs}  

\newtheorem{theorem}{\bf Theorem}[section]

\newtheorem{lemma}{\bf Lemma}[section]

\newtheorem{definition}{\bf Definition}[section]

\newtheorem{property}{\bf Property}[section]

\newtheorem{algorithm}{\bf Algorithm}

\newcommand{\bmat}{\left[ \begin{matrix}}
\newcommand{\emat}{\end{matrix} \right]}

\newcommand{\R}{\mathbb R}

\newcommand{\N}{\mathbb N}

\newcommand{\T}{\mathbb T}

\newcommand{\Ebb}{{\mathbb E}\,}

\newcommand{\cA}{\mathcal A}
\newcommand{\cB}{\mathcal B}
\newcommand{\cC}{\mathcal C}

\newcommand{\cF}{\mathcal F}
\newcommand{\cG}{\mathcal G}
\newcommand{\cH}{\mathcal H}
\newcommand{\cI}{\mathcal I}
\newcommand{\cK}{\mathcal K}

\newcommand{\cO}{\mathcal O}
\newcommand{\cP}{\mathcal P}

\newcommand{\cS}{\mathcal S}

\newcommand{\cX}{\mathcal X}

\newcommand{\bb}{\mathbf  b}

\newcommand{\bm}{\mathbf  m}

\newcommand{\bp}{\mathbf  p}
\newcommand{\bq}{\mathbf  q}

\newcommand{\bv}{\mathbf  v}
\newcommand{\bw}{\mathbf  w}
\newcommand{\bx}{\mathbf  x}
\newcommand{\by}{\mathbf  y}

\newcommand{\bA}{\mathbf  A}

\newcommand{\bC}{\mathbf  C}
\newcommand{\bD}{\mathbf  D}
\newcommand{\bE}{\mathbf  E}

\newcommand{\bI}{\mathbf  I}

\newcommand{\bM}{\mathbf  M}

\newcommand{\bP}{\mathbf  P}

\newcommand{\bS}{\mathbf  S}

\newcommand{\bW}{\mathbf  W}
\newcommand{\bX}{\mathbf  X}
\newcommand{\bY}{\mathbf  Y}

\newcommand{\bSigma}{\boldsymbol{\Sigma}}

\definecolor{darkgreen}{rgb}{0.0, 0.7, 0.0}

\graphicspath{{./figures/}}

\begin{document}

\title{Modeling and Estimation of Discrete-Time Reciprocal Processes via Probabilistic Graphical Models}
\author{Francesca Paola Carli 
\thanks{{Francesca Paola Carli is with the Department of Engineering, University of Cambridge, United Kingdom, {\tt\small fpc23@cam.ac.uk }}}
}

\markboth{DRAFT}{Shell \MakeLowercase{\textit{et al.}}: Bare Demo of IEEEtran.cls for Journals}

\maketitle

\begin{abstract} 
Reciprocal processes are acausal generalizations of Markov processes introduced  by Bernstein in 1932.
In the literature, a significant amount of attention has been focused on developing \emph{dynamical models} for reciprocal processes. 
In this paper, we provide a \emph{probabilistic graphical model} for reciprocal processes. 
This leads to a {principled solution of the smoothing problem} via \emph{message passing algorithms}. 
For the finite state space case, convergence analysis is revisited via the \emph{Hilbert metric}. 
\end{abstract}

\section{Introduction}\label{sec:Introduction}

Non causal random processes arise in many areas of science and engineering.  
For these processes, the index set usually represents space instead of time. 
The class of non causal reciprocal processes was introduced  by Bernstein in 1932 \cite{Bernstein1932} and studied by many authors \cite{Jamison1970,Jamison1974,Jamison1975,CarmichaelMasseTheodorescu1982,Krener1988, LevyFrezzaKrener1990,KrenerFrezzaLevy1991,CFPP-2011,CFPP-2013, CarravettaWhite2012, WhiteCarravetta2011}. 
A $\R^n$--valued stochastic process $\bX_{k}$ defined over the interval $\cI = [0,N]$ is said to be reciprocal if for any subinterval $[K,L]\subset \cI$, 
the process in the interior of $[K,L]$ is conditionally independent of the process in $\cI - [K,L]$ given $\bX_{K}$ and $\bX_{L}$.  
Reciprocal processes are a natural generalization of Markov processes: 
from the definition it immediately follows that Markov processes are necessarily reciprocal, but the converse is not true \cite{Jamison1970}. 
Moreover multidimensional Markov random fields reduce in one dimension to a reciprocal process, not to a Markov process. 
To attest the relevance of reciprocal processes from an engineering point of view, note, for example, that the steady-state distribution of the temperature along a heated ring or a beam subjected to random loads along its length can be modeled in terms of a reciprocal process. Applications to tracking of a ship-trajectory \cite{CastanonLevyWillsky1985}, estimation of arm movements \cite{SrinivasanEdenWillskyBrown2006}, and synthesis of textured images \cite{PicciCarli2008} have also been considered in the literature. 

Starting with Krener's work \cite{Krener1988}, a significant amount of attention has been focused on developing \emph{dynamical models} for reciprocal processes. 
Both the continuous and discrete--time case have been addressed. In this paper, our focus is on discrete--time reciprocal processes.  
In \cite{LevyFrezzaKrener1990} it has been shown that a discrete--time \emph{Gaussian} reciprocal process admits a  
second--order nearest--neighbor model driven by a locally correlated noise,  where the noise correlation structure is specified by the model dynamics. 
This model recalls state--space models for Markov processes but is \emph{acausal} (the system does not evolve recursively in the direction of increasing or decreasing values of $k$) 
and the {driving noise} is not white. 
Second order state space models for discrete--time  \emph{finite--state} reciprocal processes have been derived in \cite{CarravettaWhite2012} (see also \cite{Carravetta2008}). 

In this paper, we provide a \emph{probabilistic graphical model} for reciprocal processes with cyclic boundary conditions. 
In particular, it is shown that a reciprocal process with cyclic boundary conditions admits a single loop undirected graph as a perfect map. 
This approach is \emph{distribution--independent} and leads to a principled solution of the \emph{smoothing problem} via belief propagation (a.k.a. sum--product) algorithms.  
In this scheme, the estimated posteriors (``beliefs'') are computed as the product of incoming messages at the corresponding node,   
messages being updated through local computations (every given node updates the outgoing messages on the basis of incoming messages at the previous iteration alone). 
For tree-structured graphs, the sum--product algorithm is guaranteed to converge to the correct posterior marginal \cite{Pearl1988}. 
Nevertheless, since message passing rules are purely local, the sum--product algorithm can also be applied to loopy networks as an approximation scheme. 
As mentioned above, the graphical model associated to a reciprocal process is a single--loop network, which is not a tree. 
Convergence of sum--product algorithms for single--loop networks has been studied in the literature (see \cite{Weiss2000,WeissFreeman2001} and references therein). 
For the finite state space case, we revisit convergence analysis via the Hilbert metric. 
This approach is geometric in nature, leveraging on contraction properties of positive operators that map a quite generic cone into itself, and 
as such it can be extended to analyze convergence   
of message passing algorithms in more general settings (state--spaces, see the companion paper \cite{Carli2016-2}, and graph topologies), 
thus providing a unifying framework for the analysis of convergence of message passing algorithms 
for a single loop undirected graph,  
that has instead  been treated via ad hoc arguments in the literature 
(see e.g. \cite{Weiss2000,WeissFreeman2001} where different techniques has been employed for the Gaussian and the finite state space cases). 
 To recap, the  contribution of the paper is threefold: 
 (i) providing a probabilistic graphical model for reciprocal processes; 
 (ii) solving the smoothing problem via message passing algorithms; 
 (iii) providing an alternative analysis of convergence of such algorithms leveraging on contraction properties of positive operators with respect to the Hilbert metric.

The paper is organized as follows. 
In Section \ref{sec:RPs} reciprocal processes are introduced.  
Second--order nearest neighbor models for reciprocal processes are discussed in Section \ref{sec:SecondOrderModels-for-RPs}. 
Section \ref{sec:probabilistic-graphical-models} reviews relevant theory about probabilistic graphical models. 
The probabilistic graphical model associated to a reciprocal process with cyclic boundary conditions is derived in Section \ref{sec:PGMforRP} 
where it is shown that a reciprocal process with cyclic boundary conditions admits a single--loop Markov network as perfect-map. 
The smoothing problem for reciprocal processes is solved in Section \ref{sec:smoothing-RPs-via-BP} via loopy belief propagation. 
Sections \ref{sec:Hilbert_metric} and \ref{sec:positive-systems} introduce the Hilbert metric and discuss its relevance for stability analysis of linear positive systems. 
Contraction properties of positive operators with respect to the Hilbert metric are exploited to prove convergence of loopy belief propagation for finite state reciprocal processes in Section \ref{sec:Convergence_loopy_BP-for-RPs}. 
Section \ref{sec:Conclusions} ends the paper.

\section{Reciprocal Processes}\label{sec:RPs}

A stochastic process $\bX_{t}$ defined on a time interval $\cI$ is said to be \emph{Markov} if, for any $t_0 \in \cI$, 
the past and the future (with respect to $t_0$)  
are conditionally independent given $\bX_{t_0}$. 
A process is said to be \emph{reciprocal} if, {for each interval $[t_0, t_1] \subset \cI $}, the process in the interior of $[t_{0},t_{1}]$ and the process in $\cI - [t_{0},t_{1}] $ 
are conditionally independent given $\bX_{t_0}$ and $\bX_{t_1}$. 
Formally \cite{Jamison1974}
\begin{definition}
A  $(\cX, \Sigma)$--valued stochastic process  $\left\{ \bX_t \right\}$ on the interval $\cI$ 
with underlying probability space $\left(\Omega, \cA,P \right)$ is said to be  \emph{reciprocal} if 
\begin{equation}\label{eqn:CIs-RP}
P(AB \mid \bX_{t_{0}}, \bX_{t_{1}}) = P(A \mid \bX_{t_{0}}, \bX_{t_{1}})P(B \mid \bX_{t_{0}}, \bX_{t_{1}}), 
\end{equation}
$\forall t_{0 } < t_{1}$, $[t_{0},t_{1}] \subset \cI$, where  $A$ is the $\sigma$--field generated by the random variables $\left\{ \bX_r:  r \notin [t_{0},t_{1}] \right\}$ 
and $B$ is the $\sigma$-field generated by $\left\{ \bX_r:   r \in (t_{0},t_{1}) \right\}$.  
\end{definition} 

From the definition we have that Markov processes are necessarily reciprocal, while the converse is generally  not true \cite{Jamison1970}.
The class of reciprocal processes is thus larger than the Markov class, and it naturally extends to the multidimensional case where the parameter set of the process is not linearly ordered. 
In fact multidimensional Markov random fields reduce in one dimension to a reciprocal process, not to a Markov process. 

In this paper, we consider reciprocal processes defined on the discrete circle $\T$ with $N+1$ elements $\left\{ 0, 1, \dots, N\right\}$  
(which corresponds to imposing the cyclic boundary conditions $\bX_{-1} = \bX_{N}$,  $\bX_{N+1} = \bX_{0}$, see \cite{LevyFrezzaKrener1990, Sand96} and Section \ref{sec:SecondOrderModels-for-RPs} below)
so that the additional conditional independence relations 
\begin{align*}
\bX_{0} & \Perp \left\{ \bX_{2}, \dots, \bX_{N-1} \right\} \mid \left\{\bX_{1}, \bX_{N}\right\}\,, \\
\bX_{N} &\Perp \left\{ \bX_{1}, \dots, \bX_{N-2} \right\} \mid \left\{\bX_{0}, \bX_{N-1}\right\}
\end{align*}
hold. 

Starting with Krener's work \cite{Krener1988}, a significant amount of attention has been focused on developing \emph{dynamical models} for reciprocal processes. 
In this paper, our focus is on discrete--time reciprocal processes. 
In the next Section we briefly review dynamical models for discrete--time reciprocal processes, that were first introduced in \cite{LevyFrezzaKrener1990}. 
In Section \ref{sec:PGMforRP} we provide a \emph{probabilistic graphical model} representation of reciprocal processes.

\section{Second--order Models of Reciprocal Processes}\label{sec:SecondOrderModels-for-RPs}

Let $\bX_{k}$ be a zero-mean process defined over the finite interval $\cI=[0,N]$ and taking values in $\R^n$.  
It is well--known that if $\bX_{k}$ satisfies the recursion equation
\begin{equation}\label{eqn:markov-model}
\bX_{k+1}= \bA_{k} \bX_{k} + \bW_{k} 
\end{equation}
where $\bW_k$ is a zero--mean random process with 
\begin{equation}\label{eqn:W-white}
\Ebb\left[ \bW_k \bW^\top_l  \right]= \bI \, \delta_{kl}
\end{equation}
and $\bX_0$ is a zero--mean random variable such that 
\begin{equation}\label{eqn:cov-W-X}
\Ebb \left[\bW_k \bX_0^\top\right] = 0 
\end{equation}
then $\bX_{k}$ is Markov.  
If $\bX_{k}$ is Gaussian, then the converse is also true, namely it can be shown (see e.g. \cite{AcknerKailath1989}) that a Gaussian process is Markov if and only if it satisfies \eqref{eqn:markov-model} 
with noise structure \eqref{eqn:W-white}, \eqref{eqn:cov-W-X}. 

For a reciprocal process, the following holds. 
Let $1 \leq k \leq N-1$, and consider the model 
\begin{equation}\label{eqn:RP-model_scalar}
-\bM_{k}^{-} \bX_{k-1} + \bM_k^{0} \bX_{k} - \bM_{k}^{+} \bX_{k+1} = \bE_{k} \,, 
\end{equation}
where $\bM_k^{0}$,  $\bM_{k}^{+}$, $\bM_{k}^{-}$ are such that  
\begin{equation}\label{eqn:selfadjoint-operator_consequences}
\bM_k^{0} = (\bM_k^{0})^{\top}, \qquad \bM_{k}^{+} = (\bM_{k+1}^{-})^\top 
\end{equation}
and the driving noise $\bE_{k}$  satisfies   
\begin{equation}\label{eqn:orthogonality_scalar}
\Ebb[\bE_{k}\bX^\top_{l}] = \bI \,\delta_{kl}
\end{equation}
and is locally correlated with covariance $\bSigma_{e}$ 
\begin{equation}\label{eqn:cov_e}
[\bSigma_{e}]_{k,l}    = 
\begin{cases}
\bM_k^{0}, & \text{ for } l=k   \\
-\bM_{k}^{+}&  \text{ for } l=k+1  \\ 
0 	 & \text{otherwise} \,.
\end{cases} 
\end{equation} 
Equations \eqref{eqn:RP-model_scalar}--\eqref{eqn:cov_e} specify a second--order nearest--neighbor model. 
The model recalls standard first--order state--space models for Markov processes 
but it is \emph{acausal} (the system does not evolve recursively in the direction of increasing or decreasing values of $k$). 
Also, the {driving noise} $\bE_{k}$ is not white, but {locally correlated}.  
Notice that, in order to completely  specify 
$\bX_{k}$ over the interval $\cI = [0,N]$, 
some boundary conditions must be provided. 
Following \cite{LevyFrezzaKrener1990}, in this paper we consider \emph{cyclic boundary conditions}, namely we assume 
\begin{equation}\label{eqn:cyclic-BC}
\bX_{-1} = \bX_{N}, \qquad \bX_{N+1} = \bX_{0}\,. 
\end{equation}
These conditions are equivalent to extending cyclically the model \eqref{eqn:RP-model_scalar} and the noise structure \eqref{eqn:orthogonality_scalar}, \eqref{eqn:cov_e} to the whole interval $\cI=[0,N]$, 
provided that, in these identities, $k-1$ and $k+1$ are defined modulo $N+1$. 
Equation \eqref{eqn:RP-model_scalar} with cyclic boundary conditions \eqref{eqn:cyclic-BC} can be written in matrix form as 
\begin{equation}\label{eqn:RP-model_matricial}
\bM \bX  = \bE
\end{equation}
where 
$$
\bX = \bmat \bX_{0}\\ \bX_{1} \\ \vdots \\ \bX_{N}\emat, \quad  
\bE = \bmat \bE_{0}\\ \bE_{1} \\ \vdots \\ \bE_{N}\emat, \quad 
$$
and matrix $\bM$ given by 
\begin{equation}\label{eqn:M}
\bM = 
{\bmat 
\bM_0^{0} & -\bM_{0}^{+} & 0 & \dots & 0 &  -\bM_{0}^{-}\\
-\bM_{1}^{-}& \bM_1^{0} &-\bM_{1}^{+} & 0 & \dots & 0 \\
\dots & & & & &  \dots \\
0& \dots & 0& -\bM_{N-1}^{-}& \bM_{N-1}^{0} & -\bM_{N-1}^{+}\\
-\bM_{N}^{+}&0& \dots & 0& -\bM_{N}^{-}& \bM_N^{0} 
\emat . }
\end{equation}

It can be shown that a process $\left\{\bX_k \right\}$ satisfying \eqref{eqn:RP-model_scalar}--\eqref{eqn:cyclic-BC} is reciprocal.  
Moreover, if the process is Gaussian, 
the converse is also true. 
To be more precise: 
\begin{theorem} \cite{LevyFrezzaKrener1990}
Let $\bX_{k}$ be a zero--mean \emph{Gaussian} process {on $\T$} whose covariance $\bSigma_{x}$ is nonsingular, i.e. $\bSigma_{x} \succ 0$. 
Then $\bX_{k}$   is \emph{reciprocal} if and only if it admits a {well--posed} second--order descriptor model of the form \eqref{eqn:RP-model_scalar}--\eqref{eqn:cyclic-BC}. 
\end{theorem}

State space modeling for \emph{finite state space}  
reciprocal processes has been separately addressed in \cite{CarravettaWhite2012} (see also \cite{Carravetta2008}).  
While different state space models have been proposed in the literature for the Gaussian \cite{LevyFrezzaKrener1990} and finite state space \cite{CarravettaWhite2012} cases,   
the probabilistic graphical model we introduce in Section \ref{sec:PGMforRP}  is distribution independent. 
The following subsection provides a characterization of Gaussian reciprocal processes in terms of the sparsity pattern of their precision matrix.

\subsection*{Characterization via Covariance Matrix}\label{sec:RP_characterization_via_covariance_matrix}

If the $\bX_{k}$'s are \emph{normally distributed}, 
an important characterization in terms of sparsity pattern of the inverse of the covariance matrix (a.k.a. the precision matrix) holds. 
To start, let's recall the following (see, e.g., \cite{Dempster1972, Lauritzen1996}):  
\begin{theorem}\label{thm:zeros-in-inverse-cov-matrix-means-cond-indep}
The $(i,j)$--th (block)--entry of the inverse covariance matrix is zero 
if and only if the $i$--th and $j$--th (vector)--components of the underlying Gaussian random vector are conditionally independent given the other (vector)--components. 
\end{theorem}
Now, it is well known that $\bSigma_{x} \succ 0$ is the covariance of a (vector--valued) Markov process if and only if $\bSigma_{x}^{-1}$ is (block) tridiagonal (see \cite{AcknerKailath1989}). 
In \cite{LevyFrezzaKrener1990} the following characterization of nonsingular Gaussian reciprocal processes on 
a finite interval was obtained.
\begin{theorem}\label{thm:characterization_RP_via_covariance} 
$\bSigma_{x} \succ 0$ is the covariance matrix of the Gaussian reciprocal process \eqref{eqn:RP-model_matricial} if and only if 
$\bSigma_{x}^{-1}$ has the block tridiagonal structure 
\begin{equation}\label{eqn:covariance-matrix-RP-cyclicBC}
\bSigma_{x}^{-1} = {\bmat 
\bM_0^{0} & -\bM_{0}^{+} & 0 & \dots & 0 &  -\bM_{0}^{-}\\
-\bM_{1}^{-}& \bM_1^{0} &-\bM_{1}^{+} & 0 & \dots & 0 \\
\dots & & & & &  \dots \\
0& \dots & 0& -\bM_{N-1}^{-}& \bM_{N-1}^{0} & -\bM_{N-1}^{+}\\
-\bM_{N}^{+}&0& \dots & 0& -\bM_{N}^{-}& \bM_N^{0} 
\emat . }
\end{equation}
\end{theorem} 
If the underlying process is wide--sense stationary, then the matrices $\left\{\bM^{0}_{k}\right\}$, $\left\{\bM^{+}_{k}\right\}$, $\left\{\bM^{-}_{k}\right\} $ do not depend on $k$ and $\bSigma_{x}^{-1}$ in \eqref{eqn:covariance-matrix-RP-cyclicBC} is  block (tridiagonal and) circulant. 

By combining the two characterizations (of reciprocal and Markov process), 
if one considers the equivalence class of reciprocal processes with dynamics \eqref{eqn:RP-model_matricial}, the subclass of Markov processes is such that 
the blocks in the upper northeast corner and lower southwest corner of $\bSigma_{x}^{-1}$ are zero, i.e.
$$
\bM_{N}^{+} = (\bM_{0}^{-})^{\top}=0\,. 
$$
Theorem \ref{thm:zeros-in-inverse-cov-matrix-means-cond-indep} together with the characterization in Theorem \ref{thm:characterization_RP_via_covariance}  
will be useful in the sequel to provide an alternative derivation in the Gaussian case of the probabilistic graphical model associated to a reciprocal process.

\section{Probabilistic Graphical Models}\label{sec:probabilistic-graphical-models} 

In this section, we briefly review some relevant theory about probabilistic graphical models 
needed in the sequel 
for the derivation of the probabilistic graphical model associated to a reciprocal process (see Section  \ref{sec:PGMforRP}). 
We refer the reader to \cite{Pearl1988,Lauritzen1996,KollerFriedman2009,Bishop2006} for a thorough treatment of the subject.   

\subsection*{Graph--related  terminology and background} 

Let  $\cG = (V,E)$ be a graph where $V$ denotes the set of vertices and $E$ denotes the set of edges. 
An edge may be directed or undirected. 
In case of a directed edge from node $i$ to node $j$, we say that $i$ is a \emph{parent} of its \emph{child} $j$.
Two nodes $i$ and $j$ are \emph{adjacent} in $\cG$ 
if the directed or undirected edge $(i,j)$ is contained in $E$. 
An \emph{undirected path} is a sequence of distinct nodes $\left\{1, \dots, m \right\}$ such that there exists a (directed  or undirected) edge for each pair of nodes $\left\{l,l+1 \right\}$ on the path. 
A graph is \emph{connected} if every pair of points is joined by a path. 
A graph is \emph{singly-connected} if there exists only one undirected path between any two nodes in the graph. 
If this is not the case, the graph is said to be \emph{multiply connected}, or \emph{loopy}. 
A \emph{(un)directed cycle} is a path such that the beginning and ending nodes on the (un)directed path are the same.

If $E$ contains only undirected edges then the graph $\cG$ is an undirected graph (UG). 
If $E$ contains only directed edges then the graph $\cG$ is a directed graph (DG).

Two important classes of graphs for modeling probability distributions that we consider in this paper are UGs and directed acyclic graphs (DAGs), namely directed graphs having no directed cycles.

A graph $\cG$ is \emph{complete} if there are edges between all pairs of nodes. 
A \emph{clique} in an undirected graph is a fully connected set of nodes. 
A \emph{maximal clique} is a clique that is not a strict subset of another clique. 
An undirected graph $\cG$ is \emph{chordal} if every cycle of length greater than three has an edge connecting nonconsecutive nodes, see e.g. \cite{Golumbic80}.  
The \emph{distance} $d(u,v)$ between two vertices $u$ and $v$ in a graph $\cG$ is the length of a shortest path between them. 
If there is no path connecting the two vertices $d(u,v) = \infty$. A shortest path between any two vertices is often called a \emph{geodesic}. 
The \emph{diameter} of a (connected) graph $G$, $d(\cG)$, is the length of any longest geodesic, i.e. $d(\cG) = {\rm max}\left\{ d(u,v): u,v \, \in \, V\right\} $.

\subsection*{Probabilistic graphical models}
Now that we have introduced some terminology about graphs, we turn to the main object of this paper, namely probabilistic graphical models. 
\emph{Probabilistic graphical models} are graph--based representations that compactly encode complex  distributions over a high-dimensional space.
In a probabilistic graphical model, each node represents a random variable and the links express probabilistic relationships between these variables.
There are  different types of graphical models. 
Two major classes are \emph{Bayesian networks}, that use {directed graphs},  
and \emph{Markov networks} that are based on  {undirected graphs}. 
A third class are \emph{factor graphs}.  

There are two ways of defining a graphical model: 
(i) as a representation of a set of independencies, and  
(ii) as a skeleton for factorizing a distribution. 
For Bayesian networks, the two definitions are equivalent, while for Markov networks additional assumptions, such as having a positive distribution, 
are needed to get factorization from independencies (Hammersely--Clifford theorem). 
The primary definition of Markov networks will thus be in terms of (global) conditional independencies. 

The two formalisms, Bayesian and Markov networks,  can express different sets of conditional independencies and factorizations, 
and one or the other may be more appropriate, or even the only suitable, for a particular application. 
This will be discussed to some extent in the following. 
In this paper, we will be mainly interested in undirected graphical models, directed ones being mainly useful for expressing causal relationships between random variables. 

In this Section, we first briefly introduce Markov and Bayesian networks and then describe relevant theory that allows one to go from a given set of conditional independencies (a distribution) 
to its graph representation. 
The graphical model associated to a reciprocal process will be derived in Section  \ref{sec:PGMforRP}.

\subsection*{Markov Networks} 

The semantic of undirected graphical models is as follows. 
\paragraph{Conditional independence property} An undirected graph defines a family of probability distributions which satisfy the following graph separation property. 
\begin{property}[\textbf{Graph separation property}] \label{property:U-graph-separation} 
Let  $A$, $B$, and  $C$ denote three disjoint sets of nodes in an undirected graphical model $\cH$ and let us denote by $\bX_{A}$, $\bX_{B}$ and $\bX_{C}$ the corresponding variables in the 
associated probability distribution $P$. 
Then we say that 
$\bX_{A}  \Perp \bX_{B} \mid \bX_{C}$ (in $P$) 
($\bX_{A}$ and $\bX_{B}$ are conditionally independent given $\bX_{C}$) 
whenever (in $\cH$) there is no path from a node in $A$ to a node in $B$ which does not pass through a node in $C$.
An alternative way to view this conditional independence test is as follows: remove all nodes in set C from the graph together with any edge that connects to those nodes. 
If the resulting graph decomposes into multiple connected components such that A and B belong to different components, then $\bX_{A}  \Perp \bX_{B} \mid \bX_{C}$. 
\end{property}

\begin{paragraph}{Factorization property} 
\begin{property}[\textbf{Factorization property}]
Let $\cH$ be an undirected graphical model. 
Let $C$ be a clique and let $\bX_C$ be the set of variables in that clique. Let $\cC$ denote a set of maximal cliques. 
Define the following representation of the joint distribution
\begin{equation}\label{eqn:factorization_UGM}
p(\bx) = \frac{1}{Z} \prod_{C \in \cC} \psi_C(\bx_C)
\end{equation}
where
the functions $\psi_{C}$ can be any nonnegative valued functions (i.e. do not need to sum to 1), and are sometimes referred to as \emph{potential functions} or \emph{compatibility functions} and 
$Z$, called the \emph{partition function}, is a normalization constant chosen in such a way that the probabilities corresponding to all joint assignments sum up to $1$. 
For discrete random variables, it is given by 
$$
Z = \sum_\bx \prod_{C \in \cC} \psi_C(\bx_C)\,. 
$$
If continuous variables are considered it suffices to replace the summation by an integral. 
\end{property}
\end{paragraph}

For positive distributions, the set of distributions that are consistent with the conditional independence statements that can be read from the graph using graph separation and  
the set of distributions that can be expressed as a factorization of the form \eqref{eqn:factorization_UGM} with respect to the maximal cliques of the graph are identical. 
This is the Hammersley--Clifford theorem. 

Notice that, differently to what happens for directed graphs, potential functions in undirected graphical models generally do not have a specific probabilistic interpretation as marginal or conditional distributions.  
Only in special cases, for instance when the undirected graph is constructed by starting with a directed graph, they can admit such interpretation.

\subsection*{Bayesian Networks} 

A second class of probabilistic graphical models we shall briefly touch upon are \emph{Bayesian networks}. 
The core of the Bayesian network representation are directed acyclic graphs (DAGs).  
Similarly to Markov networks, Bayesian networks can be defined both in terms of conditional  independencies and factorization properties, but 
for Bayesian networks, the two definitions are equivalent with no need of additional assumptions. 
Reading the set of  conditional independencies encoded by a Bayesian network again needs testing whether or not the paths connecting two sets of nodes are ``blocked'',  
but the definition of ``blocked'' is this time more involved than it was for undirected graphs.  
For what concerns the factorization property, 
 in a Bayesian network, factors of the induced distribution represent the conditional distribution of a given variable conditioned on its parents. 
We do not enter here in further details about Bayesian networks models since, as we shall see, reciprocal processes do not admit a directed graph representation.

\subsection*{Factor graphs}\label{subsec:factor-graphs}

A third type of probabilistic graphical models are factor graphs. 
A \emph{factor graph} $\cF$ is an undirected graph containing two types of nodes: \emph{variable nodes} and \emph{factor nodes}. 
Suppose we have a function of several variables $\bx = \left\{\bx_1, \bx_2, \dots, \bx_N\right\}$ 
and that this function factors into a product of several functions, each having some subset of $\left\{\bx_1, \bx_2, \dots, \bx_N\right\}$ as arguments 
\begin{equation}\label{eqn:fact}
g(\bx) = \prod_s f_s(\bx_s), \qquad \text{$\bx_s$ subset of variables}
\end{equation}
This function can be represented by a factor graph having  a {variable node} for each variable $\bx_i$, 
a {factor node} (depicted by small squares) for each local function $f_s$ 
and an {edge} connecting the variable node $\bx_i$ to the factor node $f_s$ if and only if $\bx_i$ is an argument of $f_s$.

Notice that every undirected graph can be represented by an equivalent factor graph. The way to do this is to create a factor graph with the same set of variable nodes, and one factor node for each maximal clique in the graph.

\subsection*{From Models to Undirected Graphs}

So far, we have been addressing the problem of associating  a distribution to a given graphical model via the set of conditional independencies/factorization properties that the graphical model encodes. 
In this paper, we are interested in finding the probabilistic graphical model associated to a reciprocal process. 
We are thus interested in the opposite question, and namely: 
given a process defined by a set of conditional independencies, find a graphical model that encodes such a set, possibly in an ``efficient'' way. 
In other words, we want to find a graphical model that encodes all and only the conditional independencies implied by the distribution that we want to represent. 
Relevant to this aim are the notions of I-map, D-map and P-map, that we are now going to introduce. 

Consider a probability distribution $P$ and a graphical model $\cH$. 
Let $CI(P)$ denote the set of conditional independencies satisfied by $P$ and let $CI(\cH)$ denote the set of all conditional independencies implied by $\cH$. 

\begin{definition}[\textbf{I--map, D--map, P--map}]\label{def:I-D-P-maps} 
We say that 
\begin{itemize}
	\item $\cH$ is an \emph{independence map} (\emph{I--map}) for $P$ if $CI(\cH) \subset CI(P)$;
	\item $\cH$ is a \emph{dependence map}  (\emph{D--map}) for $P$ if $CI(\cH) \supset CI(P)$; 
	\item $\cH$ is a \emph{perfect map} (\emph{P-map}) for $P$ if $CI(\cH) = CI(P)$.  
\end{itemize}
\end{definition}
In other words, if $\cH$ is an I--map for $P$, then every conditional independence statement  implied by $\cH$ is satisfied by $P$. 
If $\cH$ is a D--map for $P$ then every conditional independence statement satisfied by $P$  is reflected by $\cH$. 
If it is the case that every conditional independence property of the distribution is reflected in the graph, and vice versa, then the graph is said to be a perfect map for that distribution. 
Clearly a fully connected graph will be a trivial I--map for any distribution because it implies no conditional independencies 
and a graph with no edges will be a trivial D--map for any distribution because it implies every conditional independence.

\subsection*{Generating Minimal I--maps}

Back to our original question, we have that an approach to finding a graph that represents a distribution $P$ is simply to take any graph that is an I--map for $P$. 
Yet a complete graph is an I-map for any distribution,  but there are redundant  edges in it. 
What we are really interested in, are I--maps that represent a family of distributions in a ``minimal'' way, as specified by the following definition. 
\begin{definition}[\textbf{Minimal I--map}] 
A \emph{minimal I-map} is an I-map with the property that removing any single edge 
would cause the graph to no longer be an I-map.
\end{definition}

How can we construct a minimal I-map for a distribution P? 
Here we mention two approaches for constructing a minimal I-map, one based on the pairwise Markov independencies, and the other based on the local independencies (see \cite{Pearl1988, PearlPaz1985}). 
\begin{theorem}\label{thm:construction-minimal-I-map-via-pairwise-dependencies} Let $P$ be a positive distribution, and let $\cH=(V,E)$ be defined by introducing an edge $\left( \bX, \bY\right)$ for all $\bX$, $\bY$
 that do \emph{not} satisfy $\bX \Perp \bY \mid V - \left\{ \bX,\bY\right\}$.  Then the Markov network $\cH$ is the unique minimal I-map for $P$. 
\end{theorem}

An alternative approach that uses local independencies, is based on the notion of Markov blanket, that is defined as follows. 
\begin{definition} 
Consider a  graph $\cH =(V,E)$. 
A set $U$ is a \emph{Markov blanket} of $\bX$ in a distribution $P$ if $\bX \notin U$ and if $U$ is a minimal set of nodes such that
\begin{equation}\label{eqn:MarkovBlanket}
(\bX \Perp V - \left\{ \bX\right\} - U \mid U) \in CI(P)\,.
\end{equation}
\end{definition} 
\begin{theorem}\label{thm:construction-minimal-I-map-via-MB} Let $P$ be a positive distribution. For each node $\bX$, let $MB_P(\bX)$ be a minimal set of nodes $U$  satisfying \eqref{eqn:MarkovBlanket}. 
We define a graph $\cH$ by introducing an edge $\left( \bX,\bY\right)$ for all $\bX$ and $\bY\in MB_P(\bX)$. 
Then the Markov network $\cH$ is the unique minimal I-map for $P$.
\end{theorem}

\begin{figure*}[htbp]
\centering 
\subfloat[][] {\includegraphics[width=0.29\textwidth]{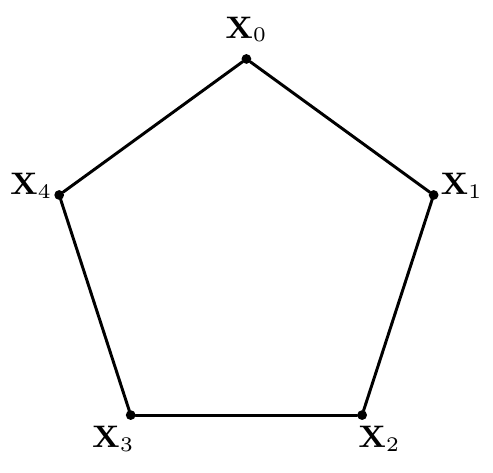}\label{fig:wrapped_timeline}}\hspace{1.5cm}
\subfloat[][] {\includegraphics[width=0.3\textwidth]{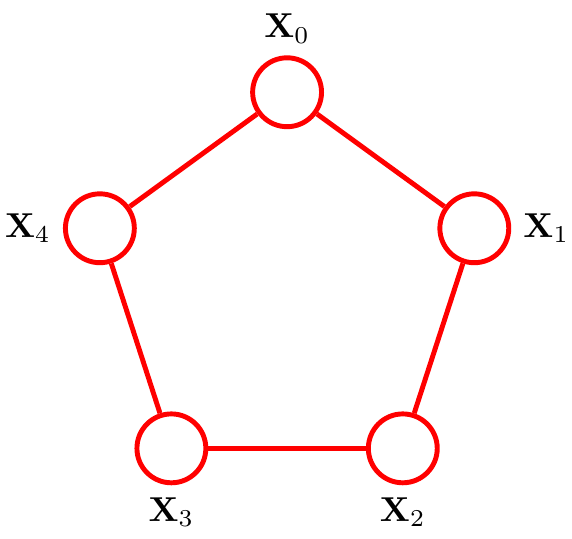}\label{fig:PGM_RP_5nodes}} 
\caption{Wrapped time line (on the left) and probabilistic graphical model (on the right) for a reciprocal process on 
{$\cI = [0,4]$}. 
\vspace{-5mm}}
\end{figure*}

\subsection*{Nonchordal Markov networks do not admit a Bayesian network as a perfect map}

It turns out that some distributions can be perfectly represented by a directed graphical model while others can be perfectly represented by an undirected one. 
On the other hand, some sets of independence assumptions can be perfectly represented both by a Bayesian network and by a Markov network. 
This is the case of undirected {chordal graphs}. The precise statement is as follows. 
\begin{theorem} \label{thm:exists-BN-as-Pmap_iff_UG-is-chordal}
Let $\cH$ be a Markov network. Then there is a Bayesian network $\cG$ such that $CI(\cH) = CI(\cG)$ if and only if $\cH$ is chordal. 
\end{theorem}

\section{Probabilistic Graphical Models of Reciprocal Processes}\label{sec:PGMforRP}

We are now ready to state our main result, namely to find the graphical model associated to a reciprocal process. 
We first derive the minimal I--map associated to a reciprocal process 
(Theorem \ref{thm:minimal-I-map-for-RP}) and then 
show that this minimal I--map is indeed also a P-map (perfect map) for the reciprocal process 
(Theorem \ref{thm:P-map-for-RP}).  

\begin{theorem} \label{thm:minimal-I-map-for-RP}
The undirected graphical model  composed of the $N+1$ nodes $\bX_{0}, \bX_{1}, \dots, \bX_{N}$ arranged in a loop (see Figure \ref{fig:PGM_RP_5nodes}) is the unique minimal I--map for a reciprocal process 
on $\cI=[0,N]$ with cyclic boundary conditions. 
\end{theorem}

\begin{IEEEproof}[Proof via Theorem \ref{thm:construction-minimal-I-map-via-MB}] 
Let $P_{R}$ denote the distribution of a reciprocal process {on $\cI$}.  
A Markov blanket of $\bX_{k}$  
in $P_{R}$ is the set $U_{R}=\left\{ \bX_{k-1}, \, \bX_{k+1} \right\}$ (where for $k=0$ and $k=N$, $k \pm 1$ has to be read modulo $N+1$). 
The undirected graphical model in Figure \ref{fig:PGM_RP_5nodes} thus follows by using the construction criterion in Theorem \ref{thm:construction-minimal-I-map-via-MB}. 
\end{IEEEproof}

For Gaussian reciprocal processes, one can also exploit the characterization in terms of sparsity pattern of the inverse covariance matrix of  Theorem \ref{thm:characterization_RP_via_covariance} and proceed as follows. 
\begin{IEEEproof}[Proof via Theorem \ref{thm:construction-minimal-I-map-via-pairwise-dependencies}] 
By Theorem \ref{thm:zeros-in-inverse-cov-matrix-means-cond-indep}, setting the $(i,j)$--th element of the inverse covariance matrix to zero has the probabilistic interpretation that the $i$--th and $j$--th components of the underlying Gaussian random vector are conditionally independent given the other components. 
The undirected graphical model in Figure \ref{fig:PGM_RP_5nodes} thus follows by the characterization of reciprocal processes in Theorem \ref{thm:characterization_RP_via_covariance}, 
using the construction criterion in Theorem \ref{thm:construction-minimal-I-map-via-pairwise-dependencies}. 
\end{IEEEproof}

\begin{theorem}\label{thm:P-map-for-RP} 
The undirected graphical model composed of the $N+1$ nodes $\bX_{0}, \bX_{1}, \dots, \bX_{N}$ arranged in a loop (see Figure \ref{fig:PGM_RP_5nodes}) is a P--map for a reciprocal process 
on $\cI=[0,N]$ with cyclic boundary conditions. 
\end{theorem}
\begin{IEEEproof} Consider a reciprocal process 
on the interval $\cI=[0,N]$ with cyclic boundary conditions $\bX_{-1}=\bX_{N}$, $\bX_{N+1}=\bX_{0}$. 
Because cyclic boundary conditions hold, one may think to the process as defined on the wrapped timeline in  Figure \ref{fig:wrapped_timeline}. 
The thesis follows from the definition of reciprocal process and the Separation Property \ref{property:U-graph-separation}
by noting that {the extremes of each} 
interval on the (wrapped) timeline define a set (pair) of nodes on the corresponding graphical model  that decomposes the graph into multiple connected components 
such that nodes corresponding to the ``interior'' and the ``exterior'' of the interval belong to different components. 
\end{IEEEproof}

\subsection*{Reciprocal processes do not admit a  directed graph as perfect map}
The Markov network in Figure \ref{fig:PGM_RP_5nodes} is \emph{not} chordal, thus, by Theorem \ref{thm:exists-BN-as-Pmap_iff_UG-is-chordal} reciprocal processes do \emph{not} admit a directed graph as perfect map. 
This is in contrast with Markov processes that admit both a directed and an undirected graphical model as perfect map.

\subsection*{Factor graph representation of a Reciprocal Process}
\begin{figure}
\begin{center}{\includegraphics[width=0.35\textwidth]{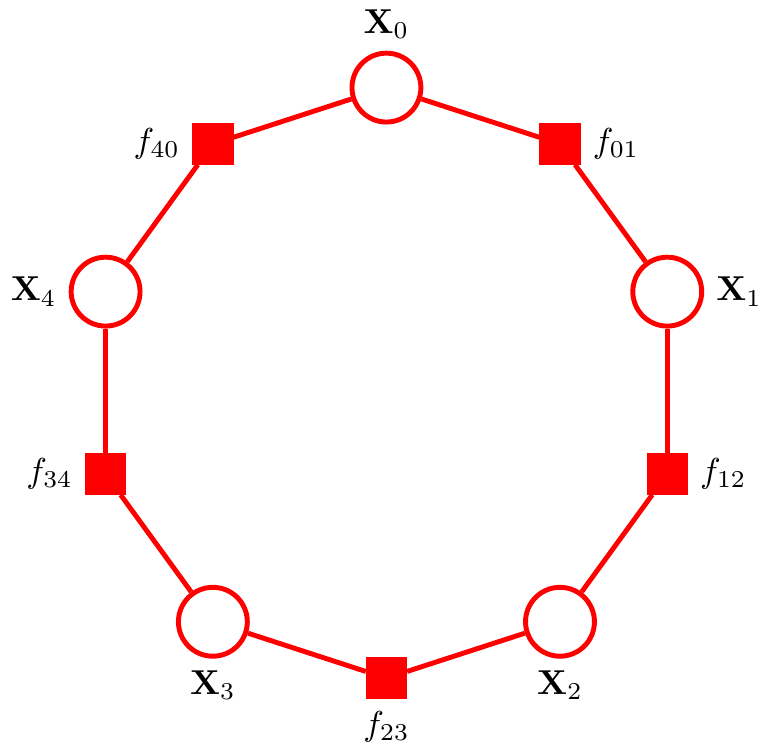}}
\caption{Factor Graph associated with a reciprocal process 
{on $\cI = [0,4]$}.  \label{fig:factor-graph-RP-5nodes} }
\end{center}
\end{figure}

As observed above, every undirected graph can be represented by an equivalent factor graph having the same set of variable nodes, and one factor node for each maximal clique in the graph. 
The factor graph corresponding to the undirected graph in Figure \ref{fig:PGM_RP_5nodes}  is shown in Figure \ref{fig:factor-graph-RP-5nodes}. 

\subsection*{Link with the four nodes  
single loop undirected graphical model by Pearl}

The single loop undirected graphical model  in Figure \ref{fig:PGM_RP_5nodes} has 
been considered in \cite[p. 90]{Pearl1988} (see also \cite{KollerFriedman2009}), 
where it has been used to model the spread of a disease 
or of a misconception among individuals who only engage in pairwise activities,  
and is used 
as a motivating example for the introduction of Markov networks, since, as observed above, the underlying set of conditional independencies does not admit a Bayesian network as a perfect map. 
Nevertheless, to the best of our knowledge, this is the first time that such graphical model is associated to a reciprocal process (distribution).  
This fills a gap in the Graphical Models literature, by bridging a well--known graphical structure 
with the class of reciprocal processes studied in the Statistics and in the Control communities. 
Moreover it opens the way to new applications 
of reciprocal processes, e.g. to the study of the spread of certain diseases or in opinion formation in social networks, that do not seem to have been explored so far 
in the literature.

\begin{figure}[htbp]
\centering 
{\includegraphics[width=0.46\textwidth]{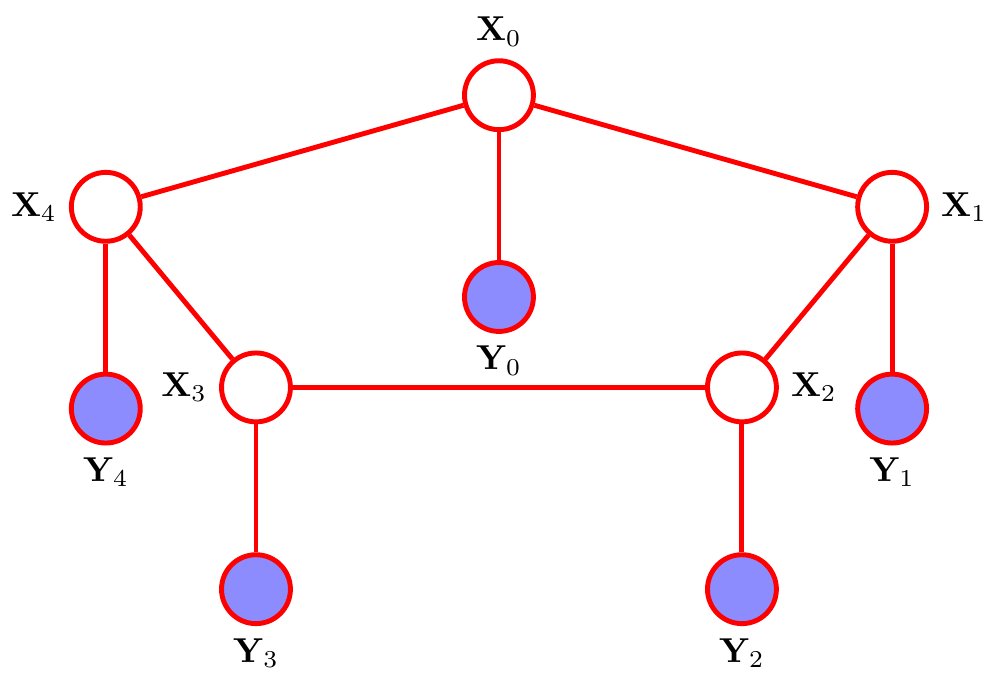}}
\caption{{Probabilistic graphical model associated to a hidden reciprocal model 
{on $\cI=[0,4]$.}}  
\label{fig:HRM_5nodes}}
\end{figure}

\section{Smoothing of Reciprocal Processes via Belief Propagation} \label{sec:smoothing-RPs-via-BP}

Consider a reciprocal process $\left\{ \bX_k\right\}$ and a second process $\left\{ \bY_k\right\}$, where, given the state sequence $\left\{ \bX_k\right\}$, the $\left\{\bY_{k}\right\}$  are independent random variables, and, for all $k$, the conditional probability distribution of $\bY_k$ depends only on $\bX_k$. In applications, $\left\{ \bX_k\right\}$ represents a ``hidden'' process which is not directly observable, 
while the observable process $\left\{ \bY_k\right\}$ represents ``noisy observations'' of the hidden process. 
We shall refer to the pair $\left\{\bX_k, \bY_{k}\right\}$ as a \emph{hidden reciprocal model}. 
The corresponding probabilistic graphical model is illustrated in Figure \ref{fig:HRM_5nodes}.  
The (fixed--interval) \emph{smoothing problem} is to compute, for all $k \in [0,N]$,  the conditional distribution of  $\bX_{k}$ given   
$\bY_{0}, \dots, \bY_{N}$. 
One of the most widespread algorithms for performing inference (solving the smoothing problem) in the graphical models literature is the \emph{belief propagation} algorithm \cite{Pearl1988, KollerFriedman2009,Bishop2006}. 
This is reviewed in Section \ref{subsec:BP} and specialized for reciprocal processes in Section \ref{subsec:BP-for-HRPs}. 
Notice that this approach is distribution independent and holds indeed both for continuous and discrete--valued  random variables/reciprocal process (even if in the Gaussian case, it may result convenient to rewrite the iteration, which lives in the infinite dimensional space of nonnegative measurable functions, in the finite dimensional spaces of 
mean vectors and  covariance matrices, see \cite{Carli2016-2} for a particularization to reciprocal processes). 
In this Section, we state the algorithm for continuous--valued variables, 
the discrete variables case following immediately by replacing integrals with summations 
where appropriate.

\subsection{Belief Propagation (a.k.a. sum--product) algorithm}\label{subsec:BP}

\begin{figure}
\begin{center}{\includegraphics[width=0.3\textwidth]{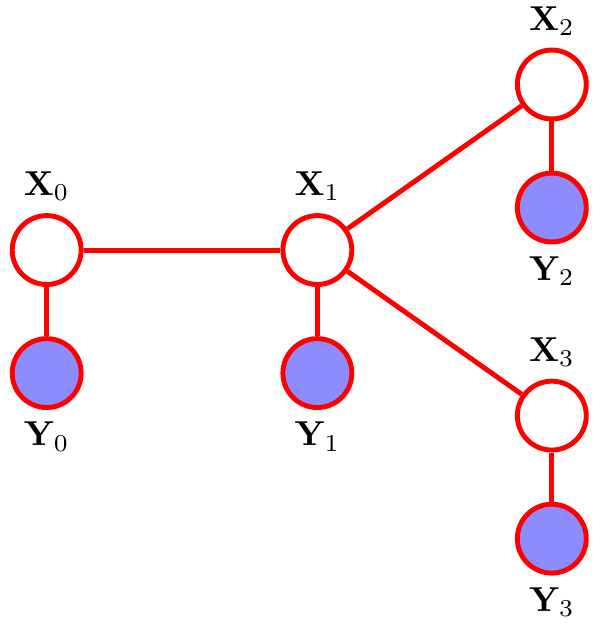}}
\caption{A graphical model with four unobserved nodes $\bX_{0}, \dots , \bX_{3}$ and four observed nodes $\bY_{0}, \dots, \bY_{3}$.  \label{fig:Vgraph}}
\end{center}
\end{figure}

Let $\cH = (E,V)$ be an undirected graphical model over the variables $\left\{ \bX_{0}, \dots, \bX_{N}\right\}$, $\bX_{i} \in  \cX$, $i=0, \dots, N$. 
In Section \ref{sec:probabilistic-graphical-models}, we have seen that the joint distribution associated with $\cH$ can be factored as 
\begin{equation}\label{eqn:factorization_UGM_2}
p(\bx) = \frac{1}{Z} \prod_{C \in \cC} \psi_C(\bx_C)\,,
\end{equation}
where $\cC$ denotes a set of maximal cliques in the graph. 
In the following, we will be interested in pairwise Markov random fields -- i.e. a Markov random field in which the joint probability factorizes into a product of bivariate potentials (potentials involving only two variables) -- where each unobserved node $\bX_{i}$ has an associated observed node $\bY_{i}$.  Factorization \eqref{eqn:factorization_UGM_2} then becomes 
\begin{equation}\label{eqn:factorization-pairwise-MRF}
p(\bx_{0:N}, \by_{0:N}) = \prod_{(i,j) \in E} \psi_{ij}(\bx_{i},\bx_{j}) \prod_{i} \psi_{i}(\bx_{i},\by_{i})\,,
\end{equation}
where the $\psi_{ij}(\bx_{i},\bx_{j})$'s are often referred to as the \emph{edge potentials} and the $\psi_{i}(\bx_{i},\by_{i})$'s are often referred to as the \emph{node potentials}. 
The problem we are interested in is finding  posterior marginals of the type  $p(\bx_{i} \mid \by_{0:N})$ for some hidden variable $\bX_{i}$. 
The basic idea behind belief propagation is to exploit the factorization properties of the distribution to allow efficient computation of the marginals.   
{Indeed, since the scope of the factors in \eqref{eqn:factorization-pairwise-MRF} is limited, this allows us to ``push in'' some of the integrals, 
performing them 
over a subset of variables at a time. }
To fix ideas, consider the graph in Figure \ref{fig:Vgraph} and suppose we want to compute the conditional marginal $p(\bx_{0}\mid \by_{0:3})$.  
A naive application of the definition, would suggest that $p(\bx_{0}\mid \by_{0:3})$ can be obtained by 
{integrating} the joint distribution over all variables except $\bX_{0}$ and then normalize 
\begin{equation}\label{eqn:brute-force_marginal_example}
p(\bx_{0} \mid \by_{0:3})\propto  \int_{\bx_1} \int_{\bx_2} \int_{\bx_3} p(\bx,\by) d\bx_{1} d\bx_{2} d\bx_{3}\,. 
\end{equation}
Nevertheless notice that the joint distribution can be factored as: 
\begin{align}\label{eqn:joint_factored_pdf_example}
p(\bx_{0:3},\by_{0:3}) =  \psi_{0}(\bx_{0})\psi_{01}(\bx_{0},\bx_{1}) &\psi_{1}(\bx_{1})  \psi_{12}(\bx_{1},\bx_{2}) \nonumber \\
&\psi_{2}(\bx_{2}) \psi_{13}(\bx_{1},\bx_{3}) \psi_{3}(\bx_{3})  \,. 
\end{align}
By plugging in factorization \eqref{eqn:joint_factored_pdf_example} into equation \eqref{eqn:brute-force_marginal_example} and  interchanging the integrals and products order, we obtain 
\begin{align}\label{eqn:marginal_x1_example}
p(\bx_{0}\mid\by_{0:3})\propto \psi_{0}(\bx_{0})\Bigg[ & \int_{\bx_{1}} \psi_{01}(\bx_{0},\bx_{1})\psi_{1}(\bx_{1}) \int_{\bx_{2}} \psi_{12}(\bx_{1},\bx_{2})\psi_{2}(\bx_{2})   \nonumber \\
&\int_{\bx_{3}} \psi_{13}(\bx_{1},\bx_{3}) \psi_{3}(\bx_{3})  \Bigg]
\end{align}
This {simple operation} forms the basis 
of the belief propagation algorithm 
and it can be given an interpretation in terms of passing of local messages around the graph. 
Most importantly, notice that it allows to reduce the computational cost of the computation of the posterior marginal from exponential to linear in the number of nodes. 
Indeed for a tree structured graph with random variables taking values in a finite alphabet $\cX$, 
the computational cost passes from $O(|\cX|^{N})$ (computational cost of the brute force marginalization in \eqref{eqn:brute-force_marginal_example}) to $O(N |\cX|^{2})$  (computational cost of the ``principled'' 
marginalization 
in \eqref{eqn:marginal_x1_example}), where $|\cX|$ denotes the cardinality of the set $\cX$.  
In its general form, the belief propagation algorithm reads as follows. 

\begin{algorithm}[Belief propagation]
Let $\bX_i$ and $\bX_j$ be two neighboring nodes in the graph. 
We denote by $m_{ij}$ the message that node $\bX_i$ sends to node $\bX_j$, by $m_{ii}$ the message that $\bY_i$ sends to $\bX_i$, 
and by $b_i$ the ``belief'' (estimated posterior marginal) at node $\bX_i$. 
The belief propagation algorithm is as follows: 
\begin{subequations}
\begin{align}
m_{ij} (\bx_{j})  &= \alpha \int_{\bx_{i}} \psi_{ij}(\bx_{i},\bx_{j}) m_{ii}(\bx_{i}) \prod_{k\in \partial i \backslash j} m_{ki} (\bx_{i})\label{eqn:message-update-scalar}\\
b_{i}(\bx_{i}) & = \beta \,\, \, m_{ii}(\bx_{i}) \prod_{k \in \partial i } m_{ki} (\bx_{i}) \label{eqn:beliefs-computation-scalar}
\end{align}
\end{subequations}
where $\partial i$ denotes the set of {neighbors} of node $\bX_{i}$ and  $\alpha$ and $\beta$ are  normalization constants so that messages and beliefs integrate to one. 
\end{algorithm}
For example, if one considers \eqref{eqn:marginal_x1_example},  
setting $m_{ii}(\bx_{i}):=  \psi_{i}(\bx_{i})$ and applying definition \eqref{eqn:message-update-scalar} for the messages, 
{taking into account that $\psi_{ij}(\bx_{i},\bx_{j})=\psi_{ji}(\bx_{j},\bx_{i})$}, \eqref{eqn:marginal_x1_example} becomes 
\begin{align*}
p(\bx_{0}\hspace{-1mm}\mid \by_{0:3})  
& \propto m_{00}(\bx_{0}) \left\{  \int_{\bx_{1}} \hspace{-1mm} \psi_{01}(\bx_{0},\bx_{1}) m_{11}(\bx_{1})   m_{21}(\bx_{1})  m_{31}(\bx_{1})  \right\} \\
& = m_{00}(\bx_{0}) \cdot m_{10}(\bx_{0})
\end{align*} 
which is of the form \eqref{eqn:beliefs-computation-scalar}, where the posterior marginal $p(\bx_{0}\mid \by_{0:3})$ is computed as the product of incoming messages at node $\bX_{0}$. 

Before going on, some observations about the belief propagation algorithm are in order. 
Observed nodes do not receive messages, and they always transmit the same message $m_{ii}$.   
The normalization of messages in equation \eqref{eqn:message-update-scalar} is not theoretically necessary (whether the messages are normalized or not, the beliefs $b_{i}$ will be identical) but helps improving numerical stability of the algorithm. 
Equation \eqref{eqn:message-update-scalar} does not specify the order in which the messages are updated. 
While a sequential scheduling policy is possible in tree--structured graphs, with messages sequentially computed starting from leaf nodes once  incoming messages at a given node become available, this is not viable 
in a loopy network. 
In this paper, following \cite{Weiss2000}, we assume that all nodes simultaneously update their messages in parallel. 
This naturally leads to \emph{loopy belief propagation}, where the update rule \eqref{eqn:message-update-scalar} is applied to graphs that are not a tree. 
This is the case of reciprocal processes, that we are going to treat in the next section. 

\subsection{Belief Propagation for Hidden Reciprocal Models}\label{subsec:BP-for-HRPs}

If the considered graph is the single--loop hidden reciprocal {model} in Figure \ref{fig:HRM_5nodes}, 
expressions \eqref{eqn:message-update-scalar} and  \eqref{eqn:beliefs-computation-scalar} for the message and belief updates simplify, each node having only two neighbors.  
Moreover we can distinguish between two classes of messages, one propagating in the direction of increasing indexes (clockwise) 
and one propagating in the direction of decreasing indexes (anticlockwise) in the loop.  
The overall algorithm with parallel scheduling policy is as follows: 

\begin{algorithm}[\textbf{(Parallel) loopy belief propagation algorithm for reciprocal processes}] \label{alg:BP-hidden-reciprocal-chain}
\begin{enumerate}
	\item Initialize all messages $\bm_{ks}^{(0)}$ to some initial value $\bar{\bm}_{ks}^{(0)}$. 

	\item For $t\in \left\{0,1, \dots, t_{max} \right\}$, for all $k \in \left\{0,1, \dots, N \right\}$
	\begin{subequations}\label{eqn:forward-backward-messages_on-a-chain}
\begin{align}
m_{k-1,\,k}^{(t+1)} (\bx_{k})  &= \alpha_{f}\hspace{-1mm}  \int_{\bx_{k-1}} \hspace{-3mm}\psi_{k-1,k}(\bx_{k-1},\bx_{k}) m_{k-1,k-1}(\bx_{k-1})  m_{k-2,k-1}^{(t)} (\bx_{k-1}) \label{eqn:mess_chain_forward} \\
m_{k+1,\,k}^{(t+1)} (\bx_{k})  &= \alpha_{b}\hspace{-1mm} \int_{\bx_{k+1}}  \hspace{-3mm} \psi_{k+1,k}(\bx_{k+1},\bx_{k}) m_{k+1,k+1}(\bx_{k+1})  m_{k+2,k+1}^{(t)} (\bx_{k+1})\,.  \label{eqn:mess_chain_backward} 
\end{align}
\end{subequations}
	
	\item For each $\bX_{k}$, $k=0, \dots, N$, compute the posterior marginals 
	\begin{equation}\label{eqn:beliefs-computation-parallel}
		b_{k}(\bx_{k})   = \beta\,\, \, m_{kk}(\bx_{k}) \left[ m_{k-1,k}^{(t_{max}+1)} (\bx_{k}) \cdot  m_{k+1,k}^{(t_{max}+1)} (\bx_{k}) \right]\,.
	\end{equation}
	
\end{enumerate}
\end{algorithm}
\vspace{3mm}
For tree-structured graphs, when $t_{max}$ is larger than the diameter of the tree (the length of the longest shortest path  between any two vertices of the graph), the algorithm converges to the correct marginal. 
This is not the case for reciprocal processes, whose associated graph is the single loop network in Figure \ref{fig:HRM_5nodes}. 
Convergence of the iteration for a hidden reciprocal model 
will be discussed in Section \ref{sec:Convergence_loopy_BP-for-RPs}. 
The argument we will use is based on contraction properties of positive operators with respect to the Hilbert metric, that we are now going to introduce.

\section{Hilbert metric}\label{sec:Hilbert_metric}

The Hilbert metric was introduced in \cite{Hilbert1895} and is defined as follows.  
Let $\cB$ be a real Banach space and let $\cK$ be a closed solid cone in $\cB$,  that is a closed subset $\cK$ with the properties that 
(i) $\lambda \cK \subset \cK$ for all $\lambda \geq 0$
(ii) the interior of $\cK$, $\cK^+$, is non--empty;  
(iii) $\cK + \cK \subseteq \cK$; 
(iv) $\cK \cap - \cK = \left\{0\right\}$. 
Define the partial order
$$x\preceq y \Leftrightarrow y-x\in \cK\,,$$
and for $x, y \in \cK \backslash \left\{0\right\}$, let 
\begin{align*}
M(x, y) &:= \inf \left\{ \lambda | x -  \lambda y \preceq 0\right\}\\
m(x, y) &:= \sup \left\{\lambda |  x -  \lambda y \succeq 0 \right\}
\end{align*}
The Hilbert metric $d_{\cH}(\cdot, \cdot)$ induced by $\cK$ is defined by 
\begin{equation}\label{eqn:Hilbert_metric}
d_{\cH}\left(x,y\right) := \log\left(\frac{M(x,y)}{m(x,y)}\right), \; \;  x,y \in \cK \backslash \left\{0\right\}\,.
\end{equation}
For example, if $\cB = \R^n$ and the cone $\cK$ is the positive orthant, $\cK = \cO:= \big\{ (x_1, \dots, x_n)\,: $ $\, x_i \geq 0, \, 1 \leq i \leq n \big\}$, 
then $M(\bx,\by)=\max_{i}(x_i/y_j)$ and $m(\bx,\by)=\min_i(x_i/y_i)$ and the Hilbert metric can be expressed as 
$$
d_{\cH}(\bx,\by) = \log \frac{\max_i(x_i/y_i)}{\min_i{(x_i/y_i)}}\,.
$$
On the other hand, if $\cB =\cS := \left\{ \bX = \bX^\top \in \R^{n \times n}\right\}$ is the set of symmetric matrices and $\cK = \cP:= \left\{ \bX \succeq 0 \mid \bX \in \cS\right\}$ is the cone of positive semidefinite matrices, then for $\bX,\bY \succ 0$, $M(\bX,\bY)=\lambda_{max}\left(\bX\bY^{-1}\right)$ and $m(\bX,\bY)= \lambda_{min}\left(\bX \bY^{-1}\right)$. Hence the Hilbert metric is 
$$
d_{\cH}(\bX,\bY) = \log \frac{\lambda_{max}\left(\bX \bY^{-1}\right)}{\lambda_{min}\left(\bX \bY^{-1}\right)}\,.
$$

An important property of the Hilbert metric is the following. 
The Hilbert metric is a \emph{projective metric} on $\cK$ i.e. it is nonnegative, symmetric, it satisfies the triangle inequality and is such that, for every $x,y \in \cK$, 
$d_{\cH}(x,y)=0$ if and only if $x=\lambda y$ for  some $\lambda > 0$. It follows easily that $d_{\cH}(x,y)$ is constant on rays, that is 
\begin{equation}\label{eqn:Hilbert_metric_invariant_under_scaling}
d_{\cH}\left(\lambda x, \mu y\right) = d_{\cH}\left(x, y\right) \quad \text{for } \lambda, \mu > 0 \,. 
\end{equation}

A second relevant property is in connection with positive operators. 
In \cite{Birkhoff1957} (see also \cite{Bushell1973}) it has been shown that  linear positive operators contract the Hilbert metric. 
This can be used to provide a geometric proof of the Perron--Frobenius theorem and, in turn, to prove attractiveness properties of linear positive systems.  
This is the subject of the next section. 

\section{Positive systems}\label{sec:positive-systems}

A linear time invariant system $\bx_{k+1} = \bA \bx_{k}$ over the positive orthant $\cO$ is positive if the mapping $\bA$ takes  $\cO$ into itself. 
Positive systems have a long history in the literature, 
both because of the relevance of the property for applications (the positivity constraint arises quite naturally when modeling real systems whose state variables represent quantities that are intrinsically nonnegative, such as pressures, concentrations, population levels, etc) 
and because the property significantly restricts the behavior, as established by Perron--Frobenius theory: if the cone invariance is strict,
that is, if the boundary of the cone is eventually mapped to the interior of the cone, then the asymptotic behavior of the system lies on a one dimensional object. 
In Section \ref{subsec:contraction-Hilbet-metric}, we briefly review contraction properties of positive linear operators as derived by Birkhoff  \cite{Birkhoff1957} (see also \cite{Bushell1973}) and then show how they can be used to prove existence of a fixed point  for 
a linear time invariant positive dynamical system which is also a global attractor (Section \ref{subsec:PerronFrobeniusVector}).

\subsection{Positive linear operators contract the Hilbert metric} 
\label{subsec:contraction-Hilbet-metric}

Let $(X,d)$ be a metric space. We recall that a mapping $f:X \rightarrow X$ is called a \emph{contraction} with respect to $d$ if there exists $0 \leq K < 1 $ such that 
\begin{equation}
d(f(x),f(y) ) \leq K d(x,y), \quad \text{for all } x,y \; \in \; X\,. 
\end{equation}
A map $A$ from $\cB$ to $\cB$ is said to be \emph{non--negative} if it takes $\cK$ into itself, i.e. 
$$
A\,: \, \cK  \rightarrow \cK\,,
$$
and \emph{positive} if it takes the interior of $\cK$ into itself, i.e. 
$$
A\,: \, \cK^{+} \rightarrow \cK^{+}\,. 
$$
For a positive map define its \emph{ contraction ratio}
\begin{equation}
k(A) := \inf \left\{  \lambda \, : d(Ax,Ay)  \leq \lambda d(x, y) \; \forall x,y \in \cK^+ \right\}
\end{equation}
and \emph{projective diameter} 
\begin{equation}
\Delta(A) := \sup \left\{ d(Ax,Ay)\, : \,  x,y \in \cK^+ \right\}\,. 
\end{equation}
It is easy to show that a non--negative linear map does not expand the Hilbert metric \cite{KohlbergPratt1982}. 
In \cite{Birkhoff1957} (see also \cite{Bushell1973}), Birkhoff showed that positivity of a linear mapping implies contraction in the Hilbert metric, 
a result that  paved the way to many contraction--based results in the literature of positive operators. 
The formal statement is as follows. 
\begin{theorem}\label{thm:properties_Hilbert_metric} 
If  $x,y \in \cK$,  then the following holds 
\begin{itemize}
	\item[$(i)$] if $A$ is a non--negative linear map on $\cK$, then $d_{\cH}(Ax,Ay) \leq d_{\cH}(x,y)$,  
	i.e. the Hilbert metric contracts weakly under the action of a non--negative linear transformation. 	
	\item[$(ii)$] [Birkhoff, 1957] If $A$ is a positive linear map in $\cB$, then  
	\begin{equation}
	k(A) = \rm{tanh} \frac14 \Delta(A)\,. 
	\end{equation}
	In other words, if the diameter $ \Delta(A)$ is finite, then positivity of a mapping implies strict contraction of the Hilbert metric. 
\end{itemize}
\end{theorem}

 In the next section we will see how contraction properties of positive linear maps with respect to the Hilbert metric can be used to explain the asymptotic behavior of a positive time--invariant system.

\subsection{Asymptotic Behavior of Positive Dynamical Systems via Contraction of the Hilbert metric}\label{subsec:PerronFrobeniusVector}

Exploiting contraction properties of positive maps, in \cite{Birkhoff1957}  (see also \cite{Bushell1973}) Birkhoff  
provided an alternative proof of the the Perron--Frobenius theorem as a special case of the Banach fixed-point theorem. 
With respect to others fixed-point arguments  used to prove the Perron-Frobenius theorem (see e.g. \cite{DebreuHerstein1953}),  
this proof has the advantage that not only it yields to the existence of a positive eigenvector $\bx_{f}$, 
but also to convergence to this same eigenvector (for the latter, in the approach in \cite{DebreuHerstein1953}, 
one still needs to show how positivity implies that the eigenvalue associated with $\bx_{f}$ dominates all the other eigenvalues). 
Along the same lines, we exploit contraction properties of positive maps with respect to the Hilbert metric to prove existence of a fixed point of the projective space for a linear time--invariant positive dynamical system 
which is a global attractor for the system. 
This will be used in the next section to prove convergence of loopy belief propagation for reciprocal processes, where we will show that the underlying iteration is indeed a positive system.  

\begin{theorem} \label{thm:convergence_BP_positive-linear-system} 
Consider the dynamical system $\bx_{k+1} = \bA \bx_{k}$ with $\bA$ a {$D \times D$} matrix with non--negative entries and suppose that $\bA$ is 
such that there exists an integer $h$ such that 
$$
[\bA^{h}]_{r,s} > 0, \quad \forall \, r,\,s \in \left\{ 1,2, \dots, D\right\}\,, 
$$
(i.e. the matrix $\bA$ is primitive). 
Then there exists a unique positive eigenvector ${\bx}_{f} \in \cO^{+}$ such that for all non--negative $\bx_{0} \in \cO \backslash \left\{ 0\right\}$, $\bA^{n} \, \bx_{0}$ converges in direction to ${\bx}_{f}$, i.e. 
$$
d_{\cH}(\bA^{n} \, \bx_{0},{\bx}_{f}) \rightarrow 0 \quad \text{ as } n \rightarrow \infty 
$$
and the rate of convergence is at least linear (i.e. the error decreases exponentially). 
\end{theorem}
To prove Theorem \ref{thm:convergence_BP_positive-linear-system} we need the two following lemmas. 

\begin{lemma}\label{thm:E_complete}\cite{Bushell1973} 
Consider the cone $\cK = \cO := \left\{ (x_1, \dots, x_D)\,:\, x_i \geq 0, \, 1 \leq i \leq D \right\}$ (positive orthant) and let $U$ denote the unit sphere in $\cB=\R^{D}$. 
Then the metric space $E:=\left\{ \cO^{+} \cap U, d_{\cH}\right\}$ is complete. 
\end{lemma}

\begin{lemma}\label{thm:proj_diam_Birkhoff}\cite{Birkhoff1957}
Let $\bA=(a_{ij})$ be a {$D \times D$} matrix with $a_{ij}>0$, for all $i,j$. Then $\bA$ is a positive map with finite projective diameter given by 
$$
\Delta(\bA) = \max \left\{ \log \frac{a_{ij}a_{pq} }{ a_{iq} a_{pj}} : 1 \leq i, \, j, \, p,\, q \leq {D} \right\} < \infty\,. 
$$ 
\end{lemma}

\begin{IEEEproof}[Proof of Theorem \ref{thm:convergence_BP_positive-linear-system}]
$\bA^{h}$ is a positive linear mapping in the interior of the positive orthant $\cO^{+}$ in $\R^{{D}}$ with finite projective diameter (see Lemma \ref{thm:proj_diam_Birkhoff}). 
Then $F(\bx) := \frac{\bA^{h} \bx}{\left\| \bA^{h} \bx \right\|}$ is a map from $E:=\left\{ \cO^{+} \cap U, d_{\cH}\right\}$ into $E$  
and is the composition of a strict contraction (see Theorem \ref{thm:properties_Hilbert_metric}(ii)) and a normalizing isometry (see \eqref{eqn:Hilbert_metric_invariant_under_scaling}). 
Since the metric space $E$ is complete (Lemma \ref{thm:E_complete}),  
then, by the Banach fixed-point theorem, there exists a unique fixed point $\bar{\bx}_{f}$ in  $E$ such that $F\left( \bar{\bx}_{f}\right)= \bar{\bx}_{f}$, 
i.e. $\bar{\bx}_{f}$ is a strictly positive eigenvector of $\bA^{h}$  (and of $\bA$), associated to a positive eigenvalue,  that, 
{starting from  an arbitrary $\bar{\bx}_{0} \in E $} (indeed for every $\bar{\bx}_{0} \, \in \cO \, \cap U$) 
can be computed as the limit of the sequence $\bar{\bx}_{k} = F( \bar{\bx}_{k-1} )$ 
so that for every $\epsilon>0$, there exists a natural number $\bar k$ such that,  for all $k > \bar k$,  $d_{\cH}(\bar \bx_k, \bar{\bx}_{f}) =  d_{\cH}(F^{k}(\bar \bx_0), \bar{\bx}_{f}  ) = 
d_{\cH}\left(\frac{\bA^{kh} \bar \bx_{0}}{\left\|\bA^{kh} \bar \bx_{0} \right\|}, \bar{\bx}_{f} \right) = d_{\cH}(\bA^{kh} \bar \bx_{0}, \bar{\bx}_{f}  ) = 
d_{\cH}(\bA^{kh}  {\bx}_{0}, {\bx}_{f}  ) 
 < \epsilon$. 
That the rate of convergence is at least linear follows by the fact that $\bA^{n}$ is a contractive map. 
QED. 
\end{IEEEproof} 

We are now ready to state our third contribution, namely to revisit convergence analysis 
of loopy belief propagation for a single loop network via contraction of the Hilbert metric.  
This is the subject of the next Section.

\section{Convergence of Loopy Belief Propagation for Reciprocal Processes} \label{sec:Convergence_loopy_BP-for-RPs}

When the graph is singly connected, local propagation rules are guaranteed to converge to the correct posterior probabilities \cite{KollerFriedman2009}. 
For general graphs with loops, theoretical understanding of the performance of local propagation schemes is an active field of research (see \cite{Weiss2000, WeissFreeman2001, IhlerFisherWillsky2005, MooijKappen2007, WainwrightJordan2008, MoallemiVanRoy2010} and references therein). 
Convergence of loopy belief propagation for networks with a single loop has been studied in \cite{Weiss2000} 
where it has been shown that, for latent variables taking values in a finite alphabet, the estimated beliefs converge 
and  accuracy of the approximation has been analyzed.  
A third contribution of this paper is to provide an alternative argument to explain convergence of loopy belief propagation that leverages on contraction properties of positive operators with respect to the Hilbert metric.  
This argument is geometric in nature 
and as such can be generalized, e.g. to the Gaussian case, 
where convergence of the iteration, which is this time a nonlinear map on the cone of positive semidefinite matrices, 
can again be traced back to contraction properties of the Hilbert metric (see \cite{Carli2016-2} for details).  
The Section is organized as follows. 
In Section \ref{subsec:BPandPositiveSystems}, we show that, for a single loop network,  convergence analysis of belief propagation  essentially boils down to the analysis of the asymptotic behavior of a linear positive dynamical system. 
Leveraging on results in Section \ref{sec:positive-systems}, convergence of the update is then derived as a consequence of contraction properties of positive linear operators with respect to the Hilbert metric (Section \ref{subsec:ContractionBasedConvergenceAnalysis}). 
Necessary and sufficient conditions for convergence of loopy belief propagation borrowed from the theory of linear positive systems are introduced in Section \ref{subsec:NSCs}. 
Following \cite{Weiss2000}, accuracy of the approximated posterior is discussed in Section \ref{subsec:AccuracyOfApproximation}.

\subsection{Belief propagation and positive systems }\label{subsec:BPandPositiveSystems}
In case of finite state space, the belief propagation algorithm can be written in matrix notation as follows. 
If one denotes with $\bM_{ij}$ the transition matrix associated with  the edge potential $\psi_{ij}(\bx_{i},\bx_{j})$ 
and by $\bm_{ij}$ (resp., $\bm_{ii}$) the vector messages obtained by staking the $m_{ij}(\bx_{i})$'s (resp., the $m_{ii}(\bx_{i})$'s) for each value of $\bX_{i}$  in $\cX =  \left\{ 0, \dots, D-1\right\}$, namely 
$$
\bm_{ij}  = \bmat m_{ij}(0) \\ \vdots \\ m_{ij}(D-1)\emat \,, \qquad
\bm_{ii} = \bmat \psi_{i}(0) \\ \vdots \\ \psi_{i}(D-1)\emat \,, 
$$
terms of the form $\sum_{x_{i}} \psi_{ij}(\bx_{i},\bx_{j}) m_{ki}(\bx_{i})$  
can be expressed in matrix notation as $\bM_{ij}\bm_{ki}$.  
Moreover we denote 
$$
\bb_{i} :=\bmat b_{i}(0) \\ \vdots \\ b_{i}(D-1)\emat \,,
$$ 
and indicate by $\odot$ the Hadamard  (entrywise) product  between two vectors of the same size. 
Thus, for a reciprocal chain, the message updates \eqref{eqn:mess_chain_forward}--\eqref{eqn:mess_chain_backward} 
can be expressed in matrix notation as 
\begin{align} 
	\bm_{k-1,k}^{(t+1)} &= \alpha_{f} \,\, \bM_{k-1,k}\left(  \bm_{k-1, k-1} \odot  \bm^{(t)}_{k-2,k-1} \right) \label{eqn:message-update-matricial-parallel-forward}\\
	\bm_{k+1,k}^{(t+1)} &= \alpha_{b} \,\, \bM_{k+1,k}\left(  \bm_{k+1,k+1} \odot  \bm^{(t)}_{k+2,k+1} \right)  \label{eqn:message-update-matricial-parallel-backward}
\end{align}
and the beliefs \eqref{eqn:beliefs-computation-parallel} as 
\begin{equation}\label{eqn:beliefs-computation-matricial-parallel}
	\bb_{k}  = \beta\,\, \, \left(  \bm_{kk} \odot  \bm_{k-1,k}^{(t_{max}+1)} \odot  \bm_{k+1,k}^{(t_{max}+1)}  \right)\,. 
\end{equation} 

Now consider the hidden reciprocal model in Figure  \ref{fig:HRM_5nodes}. 
Without loss of generality, consider the belief at node $\bX_{0}$ at a certain time {$t+N+1$}, which is given by 
\begin{equation}\label{eqn:b0_RecChain}
\bb_{0}^{(t+N+1)}  = \beta\,\, \,\bm_{00} \odot \left(  \bm_{N0}^{(t+N+1)}  \odot \bm_{10}^{(t+N+1)}\right)\,. 
\end{equation}
By the message update equation \eqref{eqn:message-update-matricial-parallel-forward}, the message that $\bX_{N}$ sends to $\bX_{0}$ at time {$t+N+1$}   
depends on the message that $\bX_{N}$ received from $\bX_{N-1}$ at time {$t+N$}
\begin{equation} \label{eqn:mN0(t+N)}
\bm_{N 0}^{{(t+N+1)} }= \alpha_{f} \bM_{N0} \left( \bm_{N N} \odot \bm_{N-1, N}^{(t+N)} \right)\,.
\end{equation} 
Similarly, the message that $\bX_{N-1}$ sends to $\bX_{N}$ at time $t+N$
depends on the message that $\bX_{N-1}$ received from $\bX_{N-2}$  at time $t+N-1$
\begin{equation}\label{eqn:mN-1N(t+N-1)}
\bm_{N-1 , N}^{(t+N)} = \alpha_{f} \bM_{N-1,N} \left( \bm_{N-1, N-1} \odot \bm_{N-2, N-1}^{(t+N-1)} \right)
\end{equation} 
and so on. One can continue expressing each message in terms of the one received from the neighbor until we go back in the loop to $\bX_{0}$: 
the message that $\bX_{0}$ sends to $\bX_{1}$ is a function of the message that $\bX_{N}$ sent to $\bX_{0}$ 
\begin{equation}\label{eqn:m0N(t+1)}
\bm_{0 1}^{(t+1)} = \alpha_{f} \bM_{01} \left( \bm_{0 0} \odot \bm_{N 0}^{(t)} \right)\,. 
\end{equation}  
By putting together \eqref{eqn:mN0(t+N)}--\eqref{eqn:m0N(t+1)}, 
one gets that the message that $\bX_{N}$ sends to $\bX_{0}$ at a given time step depends on the message that $\bX_{N}$ sent to $\bX_{0}$ $N+1$ time steps ago. 
In particular, if we denote by $\bC_{N0}$ the matrix 
\begin{equation}\label{eqn:CN0}
\bC_{N0} = \bM_{N0} \bD_{N} \bM_{N-1,N} \bD_{N-1} \cdot \dots \cdot \bM_{01} \bD_{0} 
\end{equation}
where the $\bD_{i}$'s are the diagonal matrices whose diagonal elements are the entries of the constant messages $\bm_{ii}$, 
the message that  $\bX_{N}$ sends to $\bX_{0}$ satisfy the recursion 
\begin{equation}\label{eqn:dyn-sys_BP_N0}
\bm_{N 0}^{(t+N +1)}= \alpha_{f} \bC_{N0}\bm_{N 0}^{(t)}\,. 
\end{equation} 
In a similar way, we can express the message that $\bX_{1}$ sends to $\bX_{0}$ at a given time step as a function of the message that $\bX_{1}$ sent to $\bX_{0}$ $N+1$ time steps ago 
\begin{equation}\label{eqn:dyn-sys_BP_10}
\bm_{1 0}^{(t+N+1)}= \alpha_{b} \bC_{10}\bm_{1 0}^{(t)}\,, 
\end{equation} 
where the matrix $\bC_{10}$ is given by 
\begin{equation}\label{eqn:C10}
\bC_{10} = \bM_{10} \bD_{1} \bM_{21} \bD_{2} \cdot \dots \cdot \bM_{0N} \bD_{0} \,. 
\end{equation}
Furthermore, since for any two nodes $\bX_{i}$, $\bX_{j}$, $\bM_{ij} = \bM_{ji}^{\top}$, $\bC_{10}$ can be expressed in function of $\bC_{N0}$ as 
\begin{equation}
\bC_{10} = \bD_{0}^{-1} \bC_{N0}^{\top} \bD_{0}\,. 
\end{equation}

In general, we will have two kinds of messages, traveling forward and backward in the loop, that can be written as a function of the message itself (same link and same direction) $N+1$ time steps ago. 
For indices  $k \in \left\{0, 1, \dots, N\right\}$, with $k \pm 1$ defined modulo $N+1$, we have 
\begin{subequations}
\begin{align}
\bm_{k-1, k}^{(t+N +1)}&= \alpha_{f} \bC_{k-1, k}\bm_{k-1, k}^{(t)}\,, \label{eqn:dyn-sys_BP_f}\\
\bm_{k+1, k}^{(t+N+1)}&= \alpha_{b} \bC_{k+1, k}\bm_{k+1, k}^{(t)}\,,  \label{eqn:dyn-sys_BP_b}
\end{align}
\end{subequations}
with 
\begin{subequations}
\begin{align}
\bC_{k-1, k}& = \bM_{k-1, k} \bD_{k-1} \bM_{k-2,k-1} \bD_{k-2} \cdot \dots \cdot \bM_{k,k+1} \bD_{k}\,, \\
\bC_{k+1, k} &= \bM_{k+1, k} \bD_{k+1} \bM_{k+2, k+1} \bD_{k+2} \cdot \dots \cdot \bM_{k,k-1} \bD_{k} \,,
\end{align}
\end{subequations}
where 
the similarity transformation 
\begin{equation}\label{eqn:similarity_Cf_Cb}
\bC_{k+1,k} = \bD_{k}^{-1} \bC_{k-1,k}^{\top} \bD_{k}
\end{equation}
holds.

\subsection{Contraction--based Convergence Analysis}\label{subsec:ContractionBasedConvergenceAnalysis}

From \eqref{eqn:beliefs-computation-matricial-parallel} we have that $\left( \bb_{k}^{(t+n(N+1))}\right)_{n \in \N}$ converges if the  sequences $\left( \bm_{k-1,k}^{(t+n(N+1))}\right)_{n \in \N}$, $\left( \bm_{k+1,k}^{(t+n(N+1))}\right)_{n \in \N}$ in \eqref{eqn:dyn-sys_BP_f} and \eqref{eqn:dyn-sys_BP_b}  converge. 
Now recall that the $\psi_{ij}$'s and $\psi_{i}$'s are nonnegative valued functions. 
It follows that the entries of $\bC_{k-1,k}$ and $\bC_{k+1,k}$ are nonnegative. 
The following theorem is an immediate consequence of Theorem \ref{thm:convergence_BP_positive-linear-system} 
and establishes convergence of belief propagation as a consequence of contraction of the Hilbert metric under the action of a positive linear operator. 
\begin{theorem}
Consider the hidden reciprocal model in Figure \ref{fig:HRM_5nodes} with the $\bX_{k}$'s taking values in the finite alphabet $\cX = \left\{ 0, 1, \dots, D-1\right\}$ 
and denote by $\bv_f$ and $\bw_f$ the principal eigenvector of $\bC_{k-1,k}$ and $\bC_{k+1,k}$, respectively. 
If the matrices $\bC_{k-1,k}$ and $\bC_{k+1,k}$ are primitive, then the messages $\bm_{k-1,k}$ and $\bm_{k+1,k}$ in \eqref{eqn:dyn-sys_BP_f}, \eqref{eqn:dyn-sys_BP_b}  
converge  in direction to $\bv_f$ and $\bw_f$, respectively, and the belief at node $\bX_{k}$ converges to $\bb_{k} = \bm_{kk} \odot  \bv_f \odot \bw_f$. 
The rate of convergence is at least linear. 
\end{theorem}

\subsection{Necessary and sufficient conditions for convergence}\label{subsec:NSCs}

Many necessary and sufficient conditions for asymptotic stability (e.g. the Jury criterion, the Lyapunov theorem) become simpler in the case of positive systems. 
From the theory of linear positive systems (see e.g. \cite{Luenberger1979, BermanPlemmons1994, FarinaRinaldi2000}  for details), 
the following necessary and sufficient conditions for convergence of  belief propagation for reciprocal processes follow. 

\begin{theorem}
Consider the message update \eqref{eqn:dyn-sys_BP_f} and denote by $\Lambda(\bC_{k-1,k})$ the spectrum of $\bC_{k-1,k}$. 
The following are equivalent: 
\begin{itemize}

	\item[$(i)$] the iteration \eqref{eqn:dyn-sys_BP_f} converges ($|\Lambda(\bC_{k-1,k})|<1$); 

	\item[$(ii)$] all the leading principal minors of the matrix $\bI-\bC_{k-1,k}$ are positive; 

	\item[$(iii)$] the coefficients of the characteristic polynomial of $\bC_{k-1,k}-\bI$ are positive; 

	\item[$(iv)$] there exists a diagonal matrix $\bP$ with positive diagonal elements such that the matrix $\bC_{k-1,k}^\top \bP \bC_{k-1,k} - \bP$ is negative definite. 

\end{itemize}
\end{theorem}

\subsection{Accuracy of the approximation} \label{subsec:AccuracyOfApproximation}

So far, we have been dealing with \emph{convergence} of the sequence of the beliefs $\left( \bb_{k}^{(t+n(N+1))}\right)_{n \in \N}$. 
This Section is about \emph{accuracy} of the approximation.  
The relationship between the posterior probabilities estimated via the belief propagation algorithm on a single loop network and the actual posteriors has been examined in \cite{Weiss2000}, 
and the analysis is directly applicable to the case of reciprocal processes. 
In particular, it has been shown that the smaller is the ratio between the subdominant and the dominant eigenvalue, the smaller is the approximation error. 
The result is reported here for the sake of completeness. 

\begin{theorem}\label{thm:convergence-MP-finite-state-single-loop-UGM} 
\cite{Weiss2000} 
Consider the reciprocal chain in Figure  \ref{fig:HRM_5nodes} 
{with the $\bX_{k}$, $k=0, \dots, N$ taking values in  $\cX = \left\{ 0, 1, \dots, D-1\right\}$. 
Denote by $\lambda_{1}, \lambda_{2},  \dots, \lambda_{D}$ the eigenvalues of $\bC_{k-1,k}$ (equivalently, see \eqref{eqn:similarity_Cf_Cb}, of $\bC_{k+1,k}$) sorted by decreasing magnitude 
and denote by  $\bC_{k+1,k} = \bS \Lambda \bS^{-1} $ the  eigendecomposition of $\bC_{k+1,k}$. }
Then the steady-state belief $\bb_{k}$, $k=0, \dots, N$ is related to the correct posterior marginal $\bp_{k}$ by: 
	\begin{equation}\label{eqn:b0_vs_p0}
	\bb_{k} = \beta \bp_{k} + (1-\beta) \bq_{k}
	\end{equation}
	where $\beta$ is the ratio of the largest eigenvalue of $\bC_{k-1,k}$ to the sum of all eigenvalues, 
	$\beta = {\lambda_{1}}/ \left( {\sum_{j=1}^{D} \lambda_{j}}\right)$, 
	and (the $i$--th  component of the vector) $\bq_{k}$ is given by 
	$$
	\bq_{k}(i) = \frac{\sum_{j=2}^{D}  \bS(i,j) \lambda_{j} \bS^{-1}(j,i) }{\sum_{j=2}^{D}  \bS(i,j) \lambda_{j}}\,. 
	$$
\end{theorem}	
Following \cite{Weiss2000}, we note the fundamental role played by the ratio between the subdominant and the dominant eigenvalue: 
when this ratio is small, loopy belief propagation converges rapidly and the approximation error is small. 
Indeed from \eqref{eqn:b0_vs_p0} we have 
$$ 
\bp_{k} - \bb_{k} = \left( 1- \beta \right) \left( \bq_{k} + \bp_{k}\right)\,, 
$$
i.e. the error is small when the maximum eigenvalue dominates the eigenvalue spectrum. 

In \cite{Weiss2000}, local correction formulas that compute the correct posteriors on the basis of locally available information have also been provided. 
In particular, in the case of binary latent variables, it has been shown that 
$$
p_{k}(\bx_{i}) = \frac{\lambda_{1} b_{k}(\bx_{i}) + \lambda_{2} (1-b_{k}(\bx_{i}))}{\lambda_{1} + \lambda_{2}}
$$
so that 
\begin{equation}
	p_{k}(\bx_{i}) - p_{k}(\bx_{j}) = \frac{\lambda_{1} - \lambda_{2}}{\lambda_{1} + \lambda_{2}} \left( b_{k}(\bx_{i}) - b_{k}(\bx_{j})\right)
\end{equation}
i.e. $p_{k}(\bx_{i}) - p_{k}(\bx_{j})$ is positive if and only if $b_{k}(\bx_{i}) - b_{k}(\bx_{j})$ is positive, namely the calculated beliefs are guaranteed to be on the correct side of $0.5$,  
and the correct posterior at node $\bX_{k}$, $\bp_{k}$, can be obtained from the estimated belief $\bb_{k}$ as 
\begin{equation}
	\bp_{k} = \frac{1}{1+r} \bb_{k} + \frac{r}{1+r} \left( 1 - \bb_{k} \right) \,,
\end{equation}
where $r :=  \lambda_{2} / \lambda_{1} $ and all eigenvectors and eigenvalues can be calculated by performing operations on the messages that a node receives (see  \cite{Weiss2000} for details).  
We refer the reader to  \cite[Section 4.2]{Weiss2000} for a simulation example where loopy belief propagation has been applied to a single loop network (the hidden reciprocal model in Figure \ref{fig:HRM_5nodes}).

\section{Conclusions} \label{sec:Conclusions}

In this paper, we have provided a probabilistic graphical model for reciprocal processes. 
In particular, it has been shown that a reciprocal process admits a single loop undirected graph as a perfect map. 
While in the literature a significant amount of attention has been focused on developing dynamical models for reciprocal processes, 
probabilistic graphical models for reciprocal processes 
have not been considered before. 
This approach is distribution independent and  
leads to a principled solution of the smoothing problem via message passing algorithms for graphical models. 
For the finite state space case, convergence analysis has been revisited 
leveraging on contraction properties of positive operators with respect to the Hilbert metric, an argument that is geometric in nature and as such can be extended to study convergence of message passing algorithms to more general settings (state--spaces and graph topologies, see the companion paper \cite{Carli2016-2}, 
 where convergence of belief propagation for Gaussian reciprocal processes has been addressed).


\bibliographystyle{plain}
\bibliography{biblio_ModelingAndEstimatiofRPViaProbabilisticGraphicalModels}

\end{document}